\documentclass[10pt,twocolumn,letterpaper]{article}

\usepackage{iccv}
\usepackage{times}
\usepackage{epsfig}
\usepackage{graphicx}
\usepackage{amsmath}
\usepackage{amssymb}

\usepackage[pagebackref=true,breaklinks=true,letterpaper=true,colorlinks,bookmarks=false]{hyperref}
\usepackage{multirow}
\usepackage{times}
\usepackage{epsfig}
\usepackage{graphicx}
\usepackage{diagbox}
\usepackage{interval}
\usepackage{amsmath}
\usepackage{algorithm}
\usepackage{algpseudocode}
\usepackage{boldline,multirow}
\usepackage{tabu}
\usepackage[table]{xcolor}
\usepackage{enumitem}
\usepackage{amsmath, amsthm, amssymb}

\usepackage[breaklinks=true,bookmarks=false]{hyperref}

\iccvfinalcopy 

\def\iccvPaperID{28} 

\newcommand{\norm}[1]{\left\lVert#1\right\rVert}

\ificcvfinal\fi

\begin{document}

\title{Self-Supervised Learning of Depth and Motion Under Photometric Inconsistency}

\author{Tianwei Shen\\
HKUST\\
{\tt\small tshenaa@cse.ust.hk}
\and
Lei Zhou\\
HKUST\\
{\tt\small lzhouai@cse.ust.hk}
\and
Zixin Luo\\
HKUST\\
{\tt\small zluoag@cse.ust.hk}
\and
Yao Yao\\
HKUST\\
{\tt\small yyaoag@cse.ust.hk}
\and
Shiwei Li\\
HKUST\\
{\tt\small slibc@cse.ust.hk}
\and
Jiahui Zhang\\
Tsinghua University\\
{\tt\small  zjhthu@gmail.com}
\and
Tian Fang\\
Altizure\\
{\tt\small fangtian@altizure.com}
\and
Long Quan\\
HKUST\\
{\tt\small quan@cse.ust.hk}
}

\maketitle
\ificcvfinal\thispagestyle{empty}\fi

\begin{abstract}
The self-supervised learning of depth and pose from monocular sequences provides an attractive solution by using the photometric consistency of nearby frames as it depends much less on the ground-truth data. In this paper, we address the issue when previous assumptions of the self-supervised approaches are violated due to the dynamic nature of real-world scenes. Different from handling the noise as uncertainty, our key idea is to incorporate more robust geometric quantities and enforce internal consistency in the temporal image sequence. As demonstrated on commonly used benchmark datasets, the proposed method substantially improves the state-of-the-art methods on both depth and relative pose estimation for monocular image sequences, without adding inference overhead.\end{abstract}

\section{Introduction}

The joint learning of depth and relative pose from monocular videos~\cite{vijayanarasimhan2017sfm,yin2018geonet,zhou2017unsupervised} has been an active research area due to its key role in \textit{simultaneous localization and mapping} (SLAM) and \textit{visual odometry} (VO) applications. The simplicity and the unsupervised nature make itself a potential replacement for traditional approaches that involve complicated geometric computations. Given adjacent frames, this approach uses convolutional neural networks (CNNs) to jointly predict the depth map of the target image and the relative poses from the target image to its visible neighboring frames. With the predicted depth and relative poses, photometric error is minimized between the original target image and the synthesized images formed by bilinear-sampling~\cite{jaderberg2015spatial} the adjacent views.

\begin{figure}[t]
	\resizebox{0.48\textwidth}{!}{ 
		\centering 
		\includegraphics[width=\textwidth]{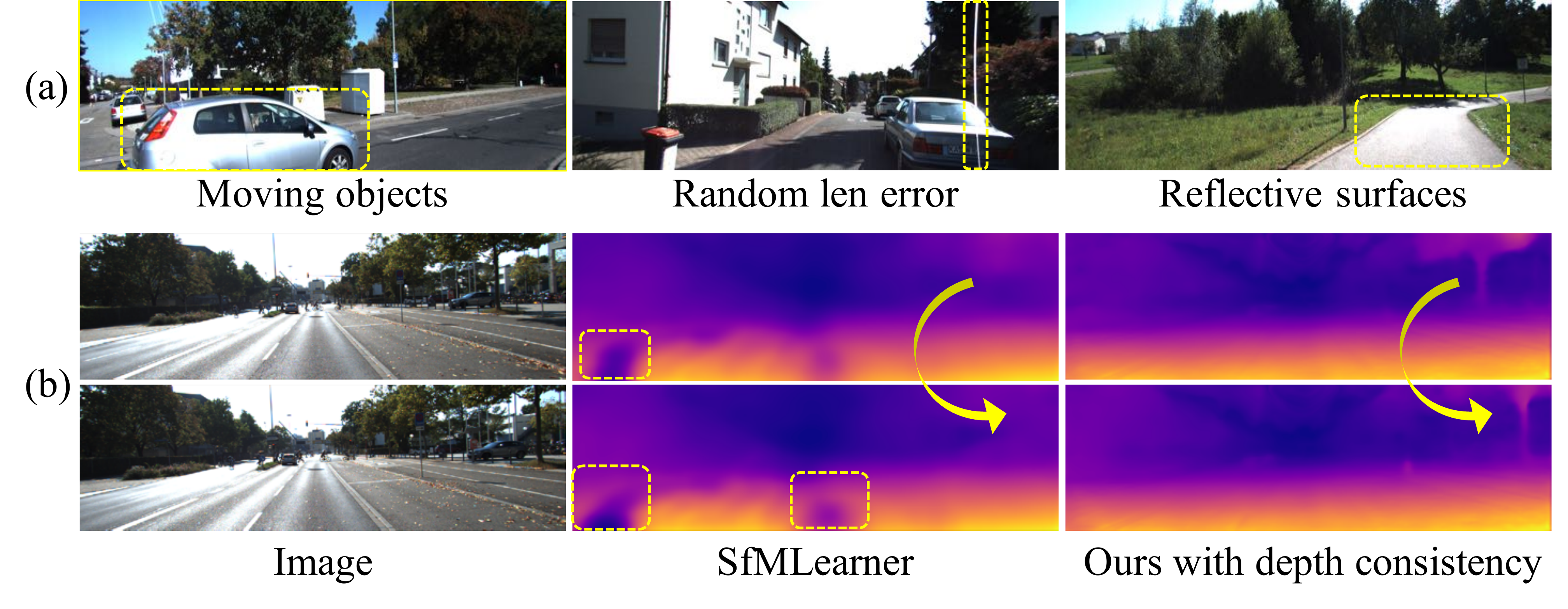}
	}
\caption{(a) Violations of the photometric consistency in KITTI~\cite{Geiger2013IJRR}. (b) The lack of depth consistency leads to erroneous `black holes' (middle~\cite{zhou2017unsupervised}) in the depth estimation.}
\label{fig:intro}

\end{figure}

However, several existing problems hinder the performance of this approach. First, the photometric loss requires the modeling scene to be static without non-Lambertian surfaces or occlusions. This assumption is often violated in street-view datasets~\cite{cordts2016cityscapes,Geiger2013IJRR} with moving cars and pedestrians (see Figure~\ref{fig:intro}(a) for some failure cases). To this end, we need other stable supervisions that are less affected when the photometric consistency is invalid. Second, as the monocular depth inference considers only single images, there is no guarantee that adjacent frames would have consistent depth estimation. This increases the chance that the inferred outcome would contain noisy depth values, and ignores the information from adjacent views when it is readily available. In addition, using pure color information is subject to the well-known gradient locality issue~\cite{bergen1992hierarchical}. When image regions with vastly different depth ranges have the similar appearance (e.g. the road in Figure~\ref{fig:intro}(b)), gradients inferred from photometric information are not able to effectively guide the optimization, leading to erroneous patterns such as `black holes' (erratic depth).

In this paper, we propose a novel formulation that emphasizes various consistency constraints of deep interplay between depth and pose, seeking to resolve the photometric inconsistency issue. We propose the geometric consistency from sparse feature matches, which is robust to illumination changes and calibration errors. We also show that enforcing the depth consistency across adjacent frames significantly improves the depth estimation with much fewer noisy pixels. The geometric information is implicitly embedded into neural networks and does not bring overhead for inference. 

\begin{figure*}[th]
	\centering 
	\includegraphics[width=\textwidth]{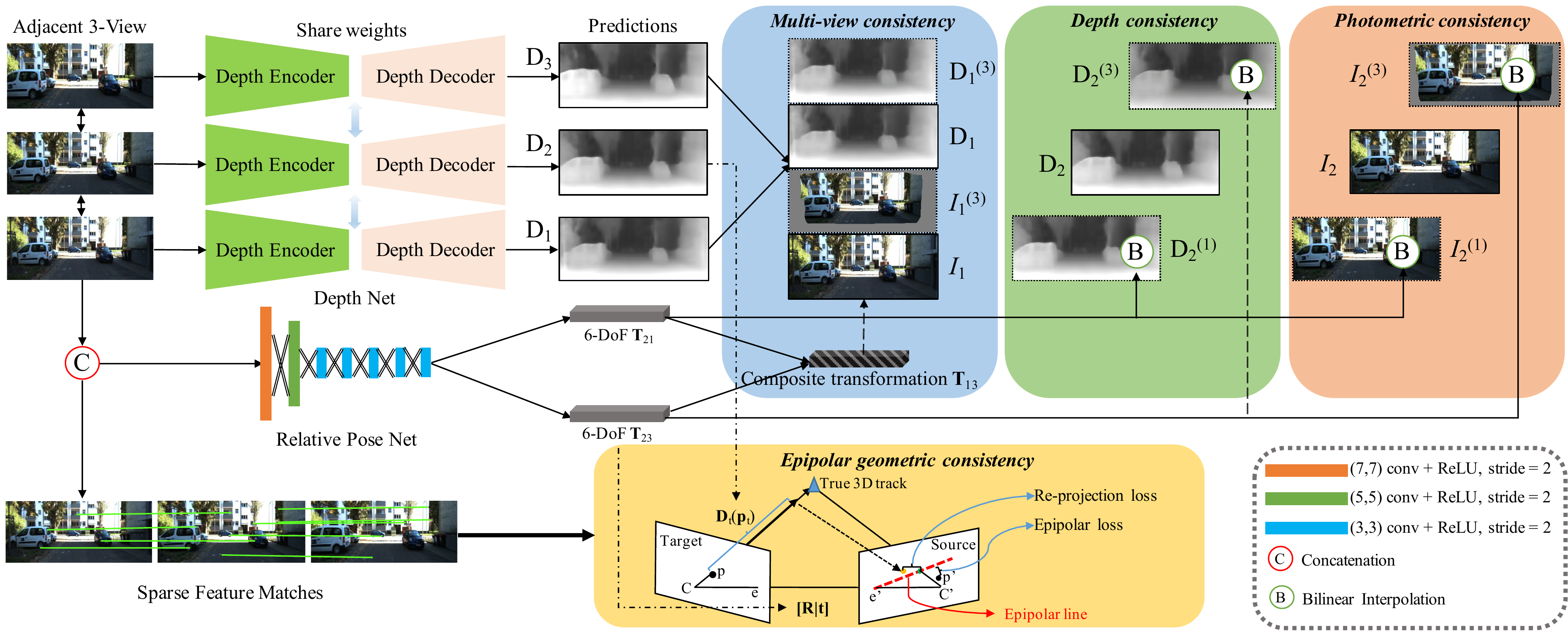}
	\caption{The architecture of our method. Besides photometric consistency (Section~\ref{sec:photometric}), we explore epipolar geometric consistency (Section~\ref{sec:epi}), depth consistency and multi-view consistency (Section~\ref{sec:depth_est}) to improve the depth and pose estimation.}
	\label{fig:arch}
\end{figure*}

The consistency of multi-view geometry has been widely applied to and even forms the basis for many sub-steps in SfM, from feature matching~\cite{baumberg2000reliable}, view graph construction~\cite{shen2016graph,zach2010disambiguating}, motion averaging~\cite{govindu2006robustness} to bundle adjustment~\cite{triggs1999bundle}.
Yet enforcing the consistency is non-trivial in the learning-based settings. Instead of tweaking the network design or learning strategy, we seek a unified framework that effectively encodes geometries in different forms, and emphasize the efficacy of geometric reasoning for the remarkable improvement. Our contributions are summarized as follows: 

\noindent(1) We introduce traditional geometric quantities based on robust local descriptors into the learning pipeline, to complement the noisy photometric loss.

\noindent(2) We propose a simple method to enforce pairwise and trinocular depth consistency in the unsupervised setting when both depth and pose are unknown.

\noindent(3) Combined with a differentiable pixel selector mask, the proposed method outperform previous methods for the joint learning of depth and motion using monocular sequences.

\section{Related Works}
\noindent\textbf{Structure-from-Motion and Visual SLAM.} 
Structure-from-Motion (SfM)~\cite{agarwal2009building} and visual SLAM problems aim to simultaneously recover the camera pose and 3D structures from images. Both problems are well studied and render practical systems~\cite{engel2017direct,mur2017orb,wu2011visualsfm} by different communities for decades, with the latter emphasizes more on the real-time performance. The self-supervised depth and motion learning framework derives from direct SLAM methods~\cite{engel2017direct,engel2014lsd,newcombe2011dtam}. Different from indirect methods~\cite{davison2007monoslam,konolige2008frameslam,mur2017orb} that use reliable sparse intermediate geometric quantities like local features~\cite{rublee2011orb}, direct methods optimize the geometry using dense pixels in the image. With accurate photometric calibration such as gamma and vignetting correction~\cite{kim2008robust}, this formulation does not rely on sparse geometric computation and is able to generate finer-grained geometry. However, this formulation is less robust than indirect ones when the photometric loss is not meaningful, the scene containing moving or non-Lambertian objects.

\smallskip\noindent\textbf{Supervised Approaches for Learning Depth.}
Some early monocular depth estimation works rely on information from depth sensors~\cite{eigen2014depth,saxena2006learning} without the aid of geometric relations. Liu \etal~\cite{liu2016learning} combine deep CNN and conditional random field for estimating single monocular images.
DeMoN~\cite{ummenhofer2017demon} is an iterative supervised approach to jointly estimate optical flow, depth and motion. This coarse-to-fine process considers the use of stereopsis and produces good results with both depth and motion supervision. 

\smallskip\noindent\textbf{Unsupervised Depth Estimation from Stereo Matching.}
Based on warping-based view synthesis~\cite{zitnick2004high}, Garg \etal~\cite{garg2016unsupervised} propose to learn depth using calibrated stereo camera pairs, in which per-pixel disparity is obtained by minimizing the image reconstruction loss. Godard \etal~\cite{godard2017unsupervised} improve this training paradigm with left-right consistency checking. Pilzer \etal~\cite{pilzer2019refine} propose knowledge distillation from cycle-inconsistency refinement. These methods use synchronized and calibrated stereo images which are less affected by occulusion and photometric inconsistency. Therefore, this task is easier than ours which uses temporal multiview images and outputs relative poses in addition.

\smallskip\noindent\textbf{Unsupervised Depth and Pose Estimation.}
The joint unsupervised optimization of depth and pose starts from Zhou~\etal~\cite{zhou2017unsupervised} and Vijayanarasimhan \etal~\cite{vijayanarasimhan2017sfm}. They propose similar approaches that use two CNNs to estimate depth and pose separately, and constrain the outcome with photometric loss. Later, a series of improvements~\cite{klodt2018supervising,mahjourian2018unsupervised,wang2018learning,yin2018geonet,zhan2018unsupervised} are proposed. Wang \etal~\cite{wang2018learning} discuss the scale ambiguity and combine the estimated depth with direct methods~\cite{steinbrucker2011real,engel2017direct}. Zhan \etal~\cite{zhan2018unsupervised} consider warping deep features from the neural nets instead of the raw pixel values. Klodt \etal~\cite{klodt2018supervising} propose to integrate weak supervision from SfM methods. Mahjourian \etal~\cite{mahjourian2018unsupervised} employ geometric constraints of the scene by enforcing an approximate ICP based matching loss. In this work, we follow the previous good practices, with the major distinction that we incorporate golden standards from indirect methods and enforce consistency terms to the state-of-the-art results.

\section{Method}
\subsection{Problem Formulation}\label{sec:photometric}
We first formalize the problem and present effective practices employed by previous methods~\cite{klodt2018supervising,mahjourian2018unsupervised,vijayanarasimhan2017sfm,wang2018learning,yin2018geonet,zhan2018unsupervised,zhou2017unsupervised}. Given adjacent $N$-view monocular image sequences (e.g. $\{\mathcal{I}_1, \mathcal{I}_2, \mathcal{I}_3 \}$ for $N=3$), the unsupervised depth and motion estimation problem aims to simultaneously estimate the depth map $\mathbf{D}_t$ of the target (center) image ($\mathcal{I}_2$ in the 3-view case) and the 6-DoF relatives poses $\mathbf{T}_{t\rightarrow s} = [\mathbf{R}_{t\rightarrow s} | \mathbf{t}_{t\rightarrow s}] \in \mathcal{SE}(3)$ to $N-1$ source views ($\mathcal{I}_1$ and $\mathcal{I}_3$), using CNNs with photometric supervision. 

For a source-target view pair $(\mathcal{I}_s, \mathcal{I}_t)$, $\mathcal{I}_t$ can be inversely warped to the source frame $\mathcal{I}_s$ given the estimated depth map $\mathbf{D}_t$ and the transformation from target to source $\mathbf{T}_{t\rightarrow s}$. Formally, given a pixel coordinate $p_t$ in $\mathcal{I}_t$ which is co-visible in $\mathcal{I}_s$, the pixel coordinate $p_s$ in $\mathcal{I}_s$ is given by the following equation which determines the warping transformation
\begin{equation}\label{eqn1}
	\small
	p_s \sim \mathbf{K}_s [\mathbf{R}_{t \rightarrow s} | \mathbf{t}_{t \rightarrow s}] \mathbf{D}_t(p_t) \mathbf{K}_t^{-1} p_t,
\end{equation}
where $\sim$ denotes `equality in the homogeneous coordinates', $\mathbf{K}_s$ and $\mathbf{K}_t$ are the intrinsics for the input image pair, and $ \mathbf{D}_t(p_t)$ is the depth for this pairticular $p_t$ in $\mathcal{I}_t$.
	
With this coordinate transformation, synthesized images can be generated from the source view using the differentiable bilinear-sampling method~\cite{jaderberg2015spatial}. The unsupervised framework then minimizes the pixel error between the target view and the synthesized image
	\begin{equation}\label{eqn:pix_loss}
	\small
	\mathcal{L}_{pixel} = \frac{1}{|\mathcal{M}|}\sum_{\forall p_t \in \mathcal{M}} \left| \widetilde{\mathcal{I}}_t^{(s)}(p_t | \mathbf{D}_t,\mathbf{T}_{t \rightarrow s}) - \mathcal{I}_t(p_t) \right|,
	\end{equation}
	where $\widetilde{\mathcal{I}}_t^{(s)}$ represents the synthesized target image from source image. $\mathcal{I}(p)$ is the function that maps the image coordinate $p$ in image $\mathcal{I}$ to pixel value, and the first term $\widetilde{\mathcal{I}}_t^{(s)}$ is the bilinear-sampling operation used to acquire the synthesized view given relative motion and depth. $\mathcal{M}$ is a binary mask that determines if the inverse warping falls into a valid region in the source image, and can be computed analytically given the per-pixel depth and relative transformation. $|\mathcal{M}|$ denotes the total number of valid pixels.
	
	In addition to the per-pixel error, structured similarity (SSIM)~\cite{wang2004image} is shown to improve the performance~\cite{godard2017unsupervised,yin2018geonet}, which is defined on local image patches $x$ and $y$ rather than every single pixel. We follow the previous approaches~\cite{mahjourian2018unsupervised,yin2018geonet} to compute the \textit{SSIM} loss on $3\times3$ image patches ($c_1 = 0.01^2, c_2 = 0.03^2$) as follows
	\begin{equation}\label{eqn:ssim}
	\small
	\mathcal{L}_{\tiny SSIM} = \frac{1}{2}[1 - \sum_{\forall x \in \widetilde{\mathcal{I}}_t^{(s)}, \forall y \in \mathcal{I}_t} \frac{(2\mu_x\mu_y+c_1)(2\sigma_{xy}+c_2)}{(\mu_x^2\mu_y^2+c_1)(\sigma_x+\sigma_y+c_2)}].
	\end{equation}
	
	The depth map is further constrained by the smoothness loss to push the gradients to propagate to nearby regions, known as the gradient locality issue~\cite{bergen1992hierarchical}. Specifically, we adopt the image-aware smoothness formulation~\cite{godard2017unsupervised,yin2018geonet} which allows sharper depth changes on edge regions
	\begin{equation}
	\small
	\mathcal{L}_{smooth} = \sum_{\forall p_t \in \mathcal{I}_t}  |\nabla \mathbf{D}_t(p_t)|^T \cdot e^{-|\nabla \mathcal{I}_t(p_t)|},
	\end{equation}
	where $\nabla$ denotes the 2D differential operator for computing image gradients. Optimizing a combination of above loss terms wraps up the basic formulation of training objectives, which forms the baseline written as
	\begin{equation}\label{eqn:base_loss}
	\small
	\mathcal{L}_{baseline} = \alpha \mathcal{L}_{pixel} + (1 - \alpha) \mathcal{L}_{\tiny SSIM} + \beta \mathcal{L}_{smooth}.
	\end{equation}
	However, there are drawbacks with the basic formulation. We then describe the key ingredients of our contributions.
	
\subsection{Learning from Indirect Methods}\label{sec:epi}
The above view synthesis formulation requires several important assumptions: 1) the modeling scene should be static without moving objects; 2) the surfaces in the scene should be Lambertian; 3) no occlusion exists between adjacent views; 4) cameras should be photometrically calibrated, a technique adopted in direct SLAM methods~\cite{engel2017direct,engel2014lsd} to compensate for vignetting~\cite{kim2008robust} and exposure time. Violation to any of the above criteria would lead to \textit{photometric inconsistency}. The first three assumptions are inevitably violated to some extent because it is hard to capture temporally static images with no occlusion in the real world. The fourth restriction is often neglected by datasets with no photometric calibration parameters provided.

To address these limitations, previous methods~\cite{klodt2018supervising,zhou2017unsupervised} additionally train a mask indicating whether the photometric loss is meaningful. Yet, we present a novel approach to tackle this issue by injecting indirect geometric information into the direct learning framework. Different from direct methods that rely on dense photometric consistency, indirect methods for SfM and visual SLAM are based on sparse local descriptors such as SIFT~\cite{wu2011visualsfm} and ORB~\cite{mur2017orb}. Local invariant features are much less likely to be affected by the scale and illumination changes and can be implicitly embedded into the learning framework.

\smallskip\noindent\textbf{Symmetric epipolar error.} Assuming the pinhole camera model, the feature matches $\mathcal{S}_{t \leftrightarrow s} = \{\mathbf{p}\leftrightarrow \mathbf{p'}\}$ between the target and source views satisfy the epipolar constraint, where $\mathbf{p}$ and $\mathbf{p'}$ are the calibrated image coordinates. 
The loss with the feature matches and the estimated pose can be quantified using the \textit{symmetric epipolar distance}~\cite{hartley2003multiple}
	\begin{equation} \label{eqn:epi}
	\small
	\begin{split}
	\mathcal{L}_{epi}( \mathcal{S} |\mathbf{R},\mathbf{t} )  = \sum_{\forall (\mathbf{p}, \mathbf{p'}) \in \mathcal{S}} & (\frac{\mathbf{p'^TEp}}{\sqrt{\mathbf{(Ep)^2_{(1)}}+\mathbf{(Ep)^2_{(2)}}}} +\\ 
	&\frac{\mathbf{p^TEp'}}{\sqrt{\mathbf{(Ep')^2_{(1)}}+\mathbf{(Ep')^2_{(2)}}}}),
	\end{split}
	\end{equation}
	where $\mathbf{E}$ being the essential matrix computed by $\mathbf{E} = [\mathbf{t}]_\times \mathbf{R}$, $[\cdot]_\times$ is the matrix representation of the cross product with $\mathbf{t}$. We simply omit the subindices for conciseness ($\mathcal{S}$ for $\mathcal{S}_{t \leftrightarrow s}$, $\mathbf{R}$ for $\mathbf{R}_{t \rightarrow s}$, $\mathbf{t}$ for $\mathbf{t}_{t \rightarrow s}$).

	\begin{figure}[th]
		\resizebox{0.48\textwidth}{!}{ 
			\centering 
			\includegraphics[width=\textwidth]{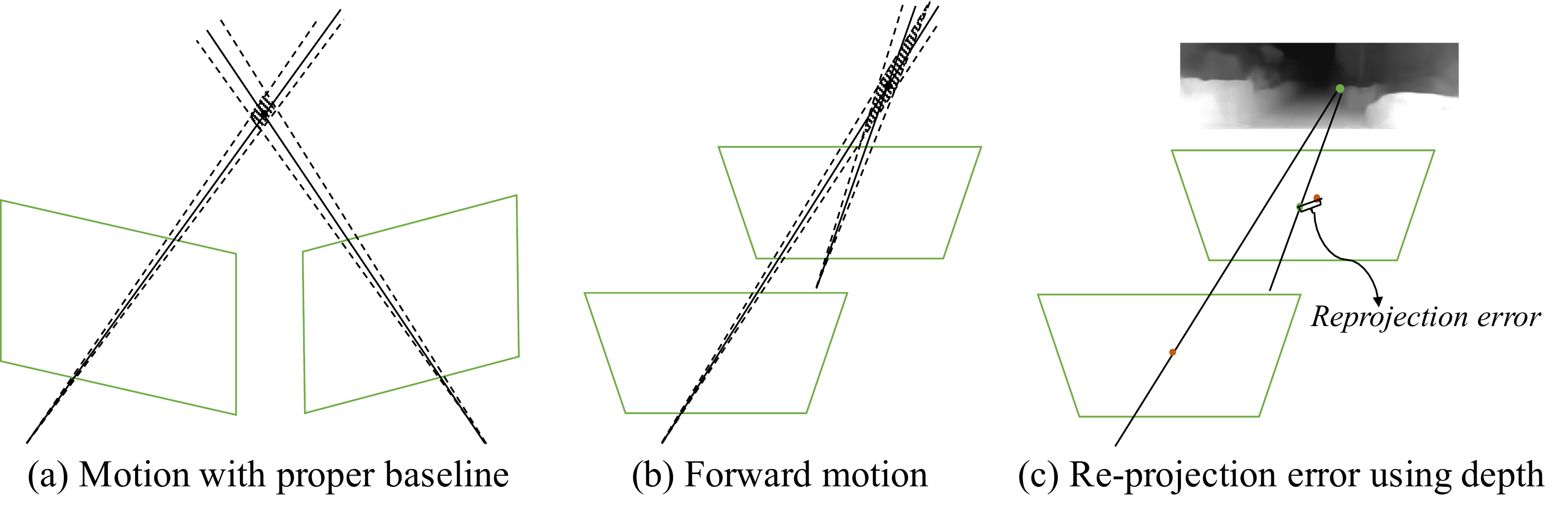}
		}
		\caption{(a) For two images with proper motion baseline, the uncertainty (shaded region) is small. (b) For forward motion with narrow baseline, the  uncertainty is large. (c) The re-projection error unites estimated depth and pose with sparse features, and does not involve triangulation uncertainty.}
		\label{fig:forward_motion}
	\end{figure}
\smallskip\noindent\textbf{Re-projection error.} The epipolar constraint does not concern the depth in its formulation. To involve depth optimization using the feature match supervision, there are generally two methods: 1) triangulate the correspondence $\mathbf{p}\leftrightarrow \mathbf{p'}$ using the optimal triangulation method~\cite{hartley2003multiple} assuming the Gaussian noise model, to obtain the 3D track for depth supervision; 2) back-project 2D features in one image using the estimated depth to compute the 3D track, and re-project the 3D track to another image to compute the re-projection error. We take the second method because the estimated depth and pose are sufficient to compute the 3D loss, and triangulation is often imprecise for ego-motion driving scenes~\cite{cordts2016cityscapes,Geiger2013IJRR} (see Figure~\ref{fig:forward_motion} for illustration, and a mathematically rigorous explanation in the Appendix). 
	\begin{equation}
	\small
	\mathcal{L}_{reproj}(\mathcal{S} | \mathbf{R}, \mathbf{t}, \mathbf{D}_t) = \sum_{\forall p\leftrightarrow p' \in \mathcal{S}} \norm{[\mathbf{R}|\mathbf{t}]\hat{\mathbf{D}}_t(\mathbf{p})\mathbf{p} - \mathbf{p}'}_2,
	\end{equation}
	where $\hat{\mathbf{D}}_t(\mathbf{p})$ is the bilinear-sampling operation~\cite{jaderberg2015spatial} in the target depth map as the feature coordinate $p$ is not an integer. Minimizing re-projection error using feature matches can be viewed as creating sparse anchors between the weak geometric supervision and the estimated depth and pose. In contrast, Equation~\ref{eqn:epi} does not involve the estimated depth.
	
Since outliers may exist if they lie close to the epipolar line, we use the pairwise matches that are confirmed in three views~\cite{hartley1997defense}. Minimizing the epipolar and re-projection errors of all matches using CNNs mimics the non-linear pose estimation~\cite{bartoli2004non}.
The experiment shows that this weak supervisory signal significantly improves the pose estimation and is superior to other SfM supervisions such as~\cite{klodt2018supervising}.

\begin{figure*}[t]
	\resizebox{\textwidth}{!}{ 
		\centering 
		\includegraphics[width=\textwidth]{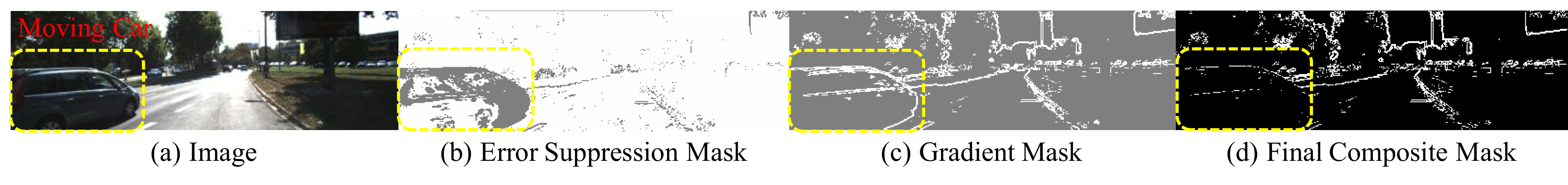}
	}
	\caption{Differentiable mask composition. The inconsistency incurred by a moving object is filtered while the meaningful loss is preserved.}
	\label{fig:final_mask}
\end{figure*}

\subsection{Consistent Depth Estimation}\label{sec:depth_est}
In this section, we describe the depth estimation module. Previous methods, whether operating on three or five views, are pairwise approaches in essence because loss terms are computed pairwisely from the source frame to the target frame. Even though the pose network outputs $N - 1$ relative poses at once, it is unknown if these relative poses are aligned to the same scale. We propose the motion-consistent depth estimation formulation to address this issue. Rather than minimizing the loss between the target frame and adjacent source frames, our proposed formulation also considers the depth and motion consistency between adjacent frames themselves.
	
\smallskip\noindent\textbf{Forward-backward consistency.} As shown in Figure~\ref{fig:arch}, our network architecture estimates the depth maps of the target image ($\mathcal{I}_t$), as well as the forward and backward depths. 
Inspired by~\cite{godard2017unsupervised,poggi2018learning} that uses left-right consistency on stereo images, we propose \textit{forward-backward consistency} for monocular images. In addition to bilinear-sampling pixel values, it samples the estimated depth maps of forward and backward images ($\mathcal{I}_s$). This process generates two synthesized depth maps $\widetilde{\mathbf{D}}_t^{(s)}$ that can be used to constrain the estimation of the target image depth map $\mathbf{D}_t$.

However, the availability of only monocular images makes the problem more challenging.
For learning with stereo images, the images are rectified in advance so the scale ambiguity issue is not considered. 
While for learning monocular depth, the estimated depth is determined only up to scale, therefore the alignment of depth scale is necessary before constraining the depth discrepancy. We first normalize the target depth map using its mean $\mathbf{D}_t : \leftarrow \mathbf{D}_t / \text{mean}(\mathbf{D}_t)$ to resolve the scale ambiguity of the target depth~\cite{wang2018learning}, which determines the scale of relative poses. Then we apply a mean alignment to the synthesized depth maps and the normalized target depth map in the corresponding region informed by the analytical mask $\mathcal{M}$ (Equation~\ref{eqn:pix_loss}), and further optimize the depth discrepancy
	\begin{equation}\label{eqn:depth_loss}
	\small
	\mathcal{L}_{depth} = \frac{1}{|\mathcal{M}|}\sum_{\forall p \in \mathcal{M}} \left|\frac{mean(\mathbf{D}_t \circ \mathcal{M})}{mean(\widetilde{\mathbf{D}}_t^{(s)} \circ \mathcal{M})} \cdot \widetilde{\mathbf{D}}_t^{(s)}(p) - \mathbf{D}_t(p)\right|,
	\end{equation}
where $\circ$ means the element-wise multiplication and the loss is averaged over all the valid pixel $p$ in the mask $\mathcal{M}$.

\smallskip\noindent\textbf{Multi-view consistency.}
The above losses are all defined on the single target image (e.g. smoothness loss) or among image pairs, even though the input is $N$-view ($N \geq 3$) image sequences. The pose network outputs $N-1$ relative poses between the target and source images, but the $N-1$ relative poses are only weakly connected by the monocular depth. To strengthen the scale consistency for triplet relation, we propose the multi-view consistency loss which penalizes inconsistency of the forward depth and backward depth using the target image as a bridge for scale alignment. Formally, given image sequence $(\mathcal{I}_1,\mathcal{I}_2,\mathcal{I}_3)$ with $\mathcal{I}_2$ the target image, and corresponding pose and depth predictions $(\textbf{T}_{2\rightarrow 1}, \textbf{T}_{2\rightarrow 3})$ and $(\mathbf{D}_1,\mathbf{D}_2,\mathbf{D}_3)$, we again obtained the normalized depth map $\overline{\mathbf{D}}_1 = s_{12} \cdot \mathbf{D}_1$ where the scaling ratio $s_{12} = \frac{mean(\mathbf{D}_2 \circ \mathcal{M}_{12})}{mean(\widetilde{\mathbf{D}}_2^{(1)} \circ \mathcal{M}_{12})}$ as used in Equation~\ref{eqn:depth_loss}. The transformation from the backward image $\mathcal{I}_1$ to the forward image $\mathcal{I}_3$ is $\mathbf{T}_{1\rightarrow 3} = \textbf{T}_{2\rightarrow 1}^{-1} \cdot \textbf{T}_{2\rightarrow 3}$. The multiview loss minimizes the depth consistency term and photometric consistency term as
\begin{equation}\label{eqn:multi}
	\small
	\begin{split}
	\mathcal{L}_{multi} =&\alpha \mathcal{L}_{pixel}(\mathcal{I}_1, \widetilde{\mathcal{I}}_1^{(3)}) + (1-\alpha) \mathcal{L}_{\tiny SSIM}(\mathcal{I}_1, \widetilde{\mathcal{I}}_1^{(3)}) \\
	+ & \frac{1}{|\mathcal{M}_{13}|}\sum_{\forall p \in \mathcal{M}_{13}} \left| \overline{\mathbf{D}}_1(p) - \overline{\mathbf{D}}_1^{(3)}(p)\right|,
	\end{split}
\end{equation}
where $\widetilde{\mathcal{I}}_1^{(3)}$ and $\overline{ \mathbf{D}}_1^{(3)}$ are the synthesized image and synthesized normalized depth given $\overline{\mathbf{D}}_3$ and $\mathbf{T}_{1\rightarrow 3}$.
The sub-indices 1 and 3 are interchangeable in the above Equation~\ref{eqn:multi}. $\mathcal{L}_{multi}$ goes beyond the pairwise loss terms $\mathcal{L}_{pixel}$, $\mathcal{L}_{SSIM}$, $\mathcal{L}_{epi}$ and $\mathcal{L}_{depth}$ because it utilizes the chained pose and pushes the two relative poses to be aligned on the same scale. This benefits monocular SLAM because it facilitates the incremental localization by aligning multiple $N$-view outputs, as we show in Section~\ref{sec:pose_est}.

\begin{table*}[]
	\centering
	\resizebox{\textwidth}{!}{ 
		\begin{tabular}{|l|c|c|c||c|c|c|c||c|c|c|}
			\hline
			Method              & Supervision   & Dataset & Cap (m) & \cellcolor{red!20}Abs Rel         &\cellcolor{red!20} Sq Rel & \cellcolor{red!20}RMSE   & \cellcolor{red!20}RMSE log &\cellcolor{green!25} $\delta <1.25 $ &\cellcolor{green!25} $\delta < 1.25^2 $& \cellcolor{green!25} $\delta < 1.25^3$            \\ \hline
			Eigen \etal~\cite{eigen2014depth} Fine   & Depth         & K       & 80      & 0.203           & 1.548  & 6.307  & 0.282    & 0.702           & 0.890           & 0.958                      \\ \hline
			Liu \etal~\cite{liu2016learning}          & Depth         & K       & 80      & 0.202           & 1.614  & 6.523  & 0.275    & 0.678           & 0.895           & 0.965                      \\ \hline
			Godard \etal~\cite{godard2017unsupervised}$\star$       & Stereo/Pose          & K       & 80      & 0.148           & 1.344  & 5.927  & 0.247    & 0.803           & 0.922           & 0.964 \\ 
			Godard \etal~\cite{godard2017unsupervised}$\star$      & Stereo/Pose          & K + CS       & 80      & \textbf{0.114}           & \textbf{0.898}  & \textbf{4.935}  & \textbf{0.206}    & \textbf{0.861}           & \textbf{0.949}       & \textbf{0.976} \\ \hline \hline
			Zhou \etal~\cite{zhou2017unsupervised} updated & No            & K       & 80      & 0.183           & 1.595  & 6.709  & 0.270    & 0.734           & 0.902           & 0.959                      \\
			Zhou \etal~\cite{zhou2017unsupervised} updated & No            & K       & -      & 0.185           & 2.170  & 6.999  & 0.271    & 0.734           & 0.901           & 0.959                      \\ \hline
			Klodt \etal~\cite{klodt2018supervising} & No            & K       & 80      & 0.166           & 1.490  & 5.998  & -    & 0.778           & 0.919           & 0.966                      \\ \hline
			Mahjourian \etal~\cite{mahjourian2018unsupervised}   & No            & K       & 80      & 0.163           & 1.24   & 6.22   & 0.25     & 0.762           & 0.916           & 0.968                      \\ \hline
			Wang \etal~\cite{wang2018learning}   & No            & K       & 80      & 0.151           & 1.257   & 5.583   & 0.228     & 0.810           & 0.936           & 0.974                      \\ \hline
			Yin \etal ~\cite{yin2018geonet}         & No            & K       & 80      & 0.155           & 1.296  & 5.857  & 0.233    & 0.793           & 0.931           & 0.973 \\ 
			Yin \etal ~\cite{yin2018geonet}         & No            & K       & -      & 0.156           & 1.470  & 6.197  & 0.235    & 0.793           & 0.931           & 0.972 \\ 
			Yin \etal ~\cite{yin2018geonet} updated        & No            & K + CS       & 80      & 0.149           & 1.060  & 5.567  & 0.226    & 0.796           & 0.935           & 0.975 \\ \hline
			Ours                & No            & K       & 80      & 0.140 & 1.025 & 5.394 & 0.222   & 0.816          & 0.938          & 0.974                     \\ 
			Ours                & No            & K       & -      & 0.140 & 1.026 & 5.397 & 0.222   & 0.816   & 0.937  & 0.974    \\ 
			Ours                & No            & K + CS       & 80      & \textbf{0.139} & \textbf{0.964} & \textbf{5.309} & \textbf{0.215}   & \textbf{0.818}          & \textbf{0.941}          & \textbf{0.977}                     \\ \hline \hline
			Garg \etal~\cite{garg2016unsupervised}         & Stereo/Pose & K       & 50      & 0.169           & 1.080  & 5.104  & 0.273    & 0.740           & 0.904           & 0.962                      \\ \hline
			Zhou \etal~\cite{zhou2017unsupervised} & No            & K       & 50      & 0.201           & 1.391  & 5.181  & 0.264    & 0.696           & 0.900           & 0.966                      \\ \hline
			Yin \etal ~\cite{yin2018geonet}         & No            & K + CS       & 50      & 0.147           & 0.936  & 4.348  & 0.218    & 0.810           & 0.941           & 0.977 \\ \hline
			Ours                & No            & K       & 50      & \textbf{0.133} & \textbf{0.778} & \textbf{4.069} & \textbf{0.207}   & \textbf{0.834}   & \textbf{0.947}  & \textbf{0.978}\emph{}    \\ \hline 
		\end{tabular}
	}
	\caption{Single-view depth estimation performance. The statistics for the compared methods are excerpted from corresponding papers, except that the results marked with \textit{`updated'} are captured from the websites. `K' represents KITTI raw dataset (Eigen split) and CS represents cityscapes training dataset. The method~\cite{godard2017unsupervised} marked with $\star$ are trained and tested on larger scale ($256\times512$) images, whereas others use $128\times416$ images. `-' in Cap(m) means no maximum depth filtering is applied. The metrics marked by {\color{red}red} means `the lower the better' and the ones marked by {\color{green}green} means `the higher the better'. The best results for each category are \textbf{bolded}.}
	\label{tab:depth_est}
\end{table*}

\subsection{Differentiable Sparse Feature Selection}
Photometric inconsistency inevitably exists due to occlusion or non-Lambertian properties. Previous works employ an additional branch to regress an uncertainty map, which helps a little~\cite{zhou2017unsupervised}. Instead, we follow the explicit occlusion modeling approach~\cite{shen2019icra} which does not rely on the data-driven uncertainty. We have observed that photometric inconsistency such as moving objects usually incurs larger photometric errors (Figure~\ref{fig:final_mask}(b)). On the other hand, image region with small gradient changes does not offer meaningful supervisory information because of the gradient locality issue~\cite{bergen1992hierarchical} (Figure~\ref{fig:final_mask}(c)). 

Therefore, we combine the error mask with the gradient mask to select the meaningful sparse features, inspired by the direct sparse odometry~\cite{engel2017direct} but can be fit into the differentiable training pipeline. Given the pixel error map, we compute the error histogram and mask out the pixels which are above $\sigma_{e}(=90)$-th percentile. We also compute the gradient mask and keep only the values that are above $\sigma_{g}(=90)$-th percentile. The final composite mask is the multiplication of both masks with dynamic thresholding. As shown in Figure~\ref{fig:final_mask}(d), this mask operation filters out a majority of photometric inconsistency regions like the moving car. The composite mask is only used for the final depth refinement when the error suppression mask is stable, otherwise we observe a performance drop if training from scratch.

Our final formulation takes into account the basic losses in Equation~\ref{eqn:base_loss}, the geometric terms, as well as the consistency terms, written as
\begin{equation} \label{eqn:total_loss}
\small
\begin{split}
\mathcal{L}_{total} = &\alpha \mathcal{L}_{pixel} + (1-\alpha)\mathcal{L}_{SSIM}+\beta \mathcal{L}_{smooth} + \\
&\gamma_1 \mathcal{L}_{epi} + \gamma_2 \mathcal{L}_{reproj} + \mu_1 \mathcal{L}_{depth} + \mu_2 \mathcal{L}_{multi}.
\end{split}
\end{equation}
The weighting for different losses are set empirically given hyper-parameters in previous methods and our attempts ($\alpha=0.15, \beta=0.1, \gamma_1=\gamma_2=0.001, \mu_1=\mu_2=0.1$). We also try to learn the optimal weighting using homoscedastic uncertainty~\cite{kendall2017geometric}, but find no better result than empirically setting the weights.

\section{Experiments}
\subsection{Training Dataset}
\noindent\textbf{KITTI.} We evaluate our method on the KITTI datasets~\cite{Geiger2013IJRR,Menze2015CVPR}, using the raw dataset with Eigen split~\cite{eigen2014depth} for depth estimation, and the odometry dataset for pose estimation. Images are down-sampled to 128 $\times$ 416 to facilitate the training and provide a fair evaluation setting. For Eigen split, we use 20129 images for training and 2214 images for validation. The 697 testing images are selected by~\cite{eigen2014depth} from 28 scenes whose images are excluded from the training set. For the KITTI odometry dataset, we follow the previous convention~\cite{yin2018geonet,zhou2017unsupervised} to train the model on sequence 00-08 and test on sequence 09-10. We further split sequence 00-08 to 18361 images for training and 2030 for validation.

\smallskip\noindent\textbf{Cityscapes.} We also try pre-training the model on the Cityscapes~\cite{cordts2016cityscapes} dataset since starting from a pre-trained model boosts the performance~\cite{zhou2017unsupervised}. The process is conducted without adding feature matches for 60k steps. 88084 images are used for training and 9659 images for validation.

\subsection{Implementation Details}
\noindent\textbf{Data preparation.} We extract SIFT~\cite{lowe2004distinctive} feature matches as the weak geometric supervision using SiftGPU~\cite{wu2011visualsfm} offline.
The putative matches are further filtered by geometric verification~\cite{hartley1997defense} with RANSAC~\cite{fischler1981random}. 100 feature pairs are randomly sampled and used for training. Matches are only used for training and not necessary for inference.

\smallskip\noindent\textbf{Learning.} We implement our pipeline using Tensorflow~\cite{abadi2016tensorflow}.
The depth estimation part follows~\cite{yin2018geonet} which uses ResNet-50~\cite{he2016deep} as the depth encoder. The relative pose net follows~\cite{zhou2017unsupervised,yin2018geonet} which is a 7-layer CNN, with the lengths of feature maps reduced by half and the number of feature channels multiplied by two from each previous layer. If not explicitly specified, we train the neural networks using 3-view image sequences as the photometric error would accumulate for longer input sequences. We use the Adam~\cite{kingma2014adam} solver with $\beta_1 = 0.9$, $\beta_2 = 0.999$, a learning rate of 0.0001 and a batch size of 4.

\smallskip\noindent\textbf{Training efficiency.} The proposed method takes longer time per step due to more depth estimations and loss computations. With a single GTX 1080 Ti, training takes 0.35s per step compared with 0.19s for the baseline approach based on Equation~\ref{eqn:base_loss}. It is noted that the inference efficiency is the same as the baseline.

\begin{figure*}[t]
	\centering
	\includegraphics[width=0.98\textwidth]{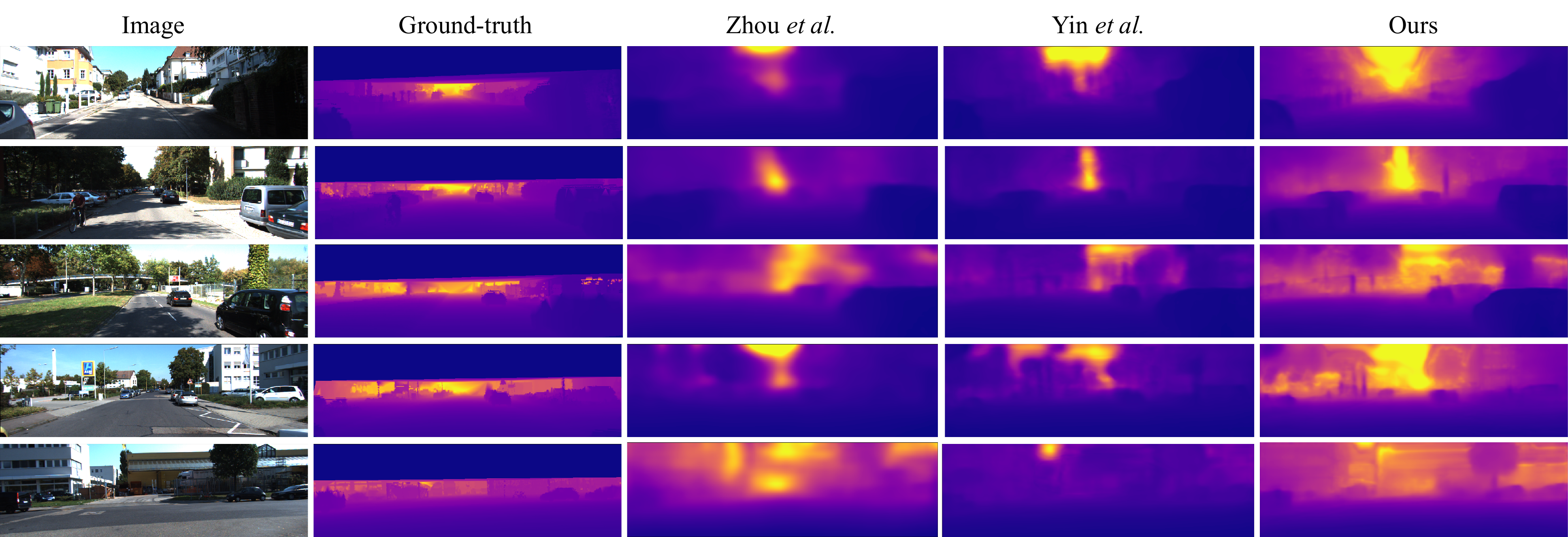}
	\caption{Qualitative comparison for depth estimation on the Eigen split. The predicted depth maps are first aligned with the ground-truth using mean. Then the depth values larger than 80m are set to 80m to ensure a consistent color scale. It shows that our result best reflects the ground-truth depth range and contains richer details (best view in color).}
	\label{fig:eigen_depth}
\end{figure*}
\subsection{Depth Estimation}
The evaluation of depth estimation follows previous works~\cite{mahjourian2018unsupervised,yin2018geonet,zhou2017unsupervised}. As shown in Table~\ref{tab:depth_est}, our method achieves the best performance among all unsupervised methods that jointly learn depth and pose. Previous methods often filter the predicted depth map by setting a maximum depth at 50m or 80m (the ground-truth depth range is within 80m) before computing depth error, since distant pixels may have prediction outliers. We also evaluate the performance without this filtering step, marked by `-' in the \textit{Cap(m)} column. It shows that without capping the maximum depth, ~\cite{yin2018geonet,zhou2017unsupervised} become worse while our result seldom changes, meaning our consistent training renders depth predictions with little noise. Figure~\ref{fig:eigen_depth} provides a qualitative comparison of the predictions. We show the depth value (the nearer the darker) instead of the inverse depth (disparity) parameterization, which highlights the distant areas.

Since both Klodt \etal~\cite{klodt2018supervising} and ours use self-supervised weak supervisions, we redo the experiments in~\cite{klodt2018supervising} that use self-generated poses and sparse depth maps from ORB-SLAM2~\cite{mur2017orb} for weak supervision, fixing other settings. We obtain slightly better statistics which still lags behind the proposed method that uses feature matches.
This implies that the raw matches are more robust as the supervisory signal, whereas using pose and depth computed from SfM/SLAM is possible to introduce additional bias inherited from the PnP~\cite{lepetit2009epnp} or triangulation algorithms.

\subsection{Pose Estimation}\label{sec:pose_est}
\begin{table}[] 
	\centering
	\resizebox{0.48\textwidth}{!}{ 
		\begin{tabular}{|l|c|c|}
			\hline
			Method                & Seq 09 & Seq 10 \\ \hline
			ORB-SLAM2~\cite{mur2017orb}             & 0.014 $\pm$ 0.008   &  0.012 $\pm$ 0.011      \\ \hline
			Zhou \etal~\cite{zhou2017unsupervised} updated (5-frame)              &  0.016 $\pm$ 0.009      & 0.013 $\pm$ 0.009       \\ \hline
			Yin \etal ~\cite{yin2018geonet} (5-frame)            &    0.012 $\pm$ 0.007     &   0.012 $\pm$ 0.009     \\ \hline
			Mahjourian \etal~\cite{mahjourian2018unsupervised} , no ICP (3-frame)      &   0.014 $\pm$ 0.010      &   0.013 $\pm$ 0.011      \\ \hline
			Mahjourian \etal~\cite{mahjourian2018unsupervised} , with ICP (3-frame)     &  0.013 $\pm$ 0.010      &  0.012 $\pm$ 0.011       \\ \hline
			Klodt \etal~\cite{klodt2018supervising} (5-frame)          &   0.014 $\pm$ 0.007     &    0.013 $\pm$ 0.009     \\ \hline
			Ours \etal (3-frame) &   \textbf{0.009 $\pm$ 0.005}     & \textbf{0.008 $\pm$ 0.007}       \\ \hline
		\end{tabular}
	}
	\caption{Pose estimation evaluation. All the learning-based methods are trained and tested on 128 $\times$ 416 images, while ORB-SLAM2 are tested on full-sized ($370\times1226$) images.}
		\label{tab:pose}
\end{table}

\begin{figure}[th]
	\resizebox{0.48\textwidth}{!}{ 
		\centering 
		\includegraphics[width=\textwidth]{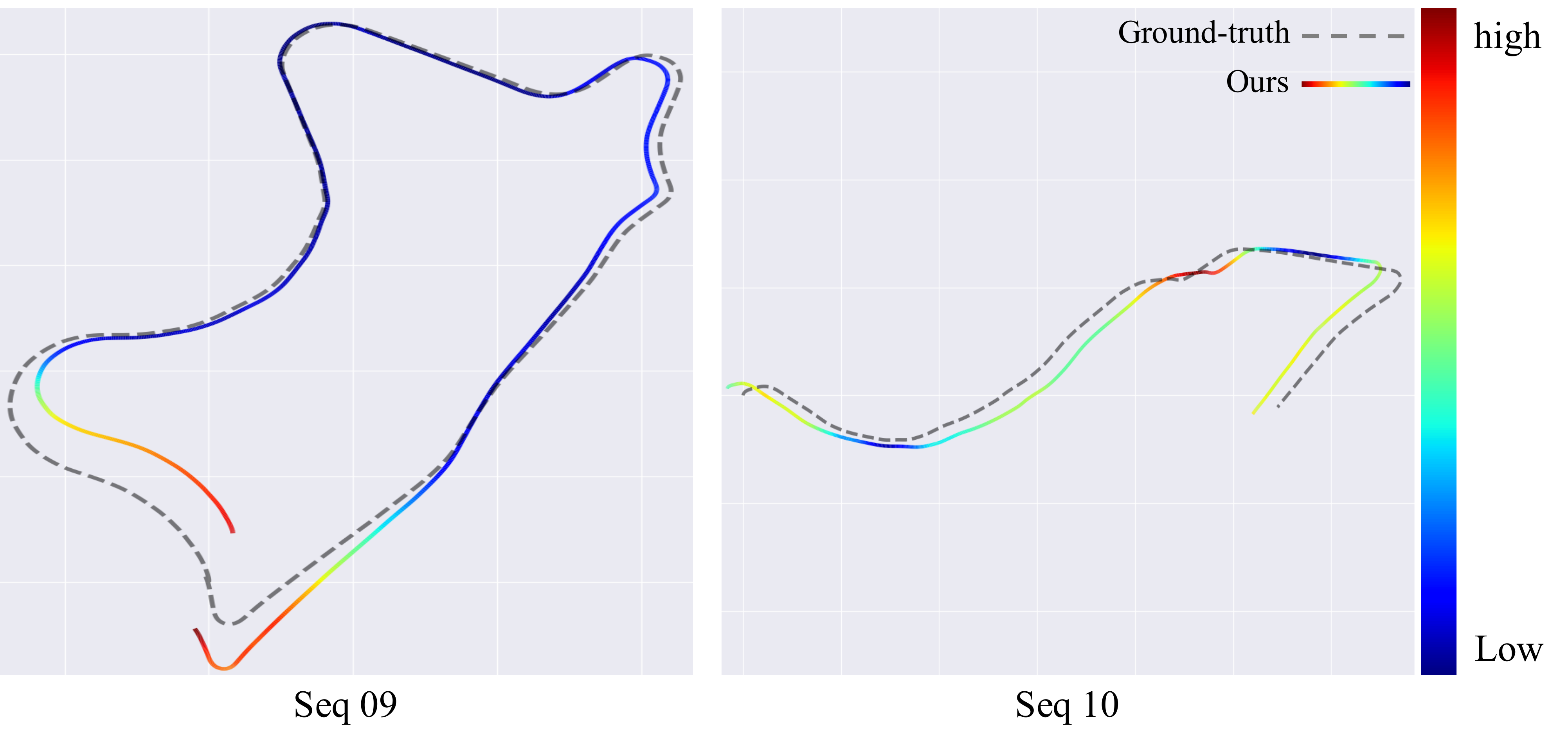}
	}
	\caption{The chained trajectory of the proposed method drawed with the ground-truth on KITTI sequence 09/10. The color bar on the right shows the scale of alignment error.}
	\label{fig:slam_traj}
\end{figure}
We evaluate the performance of relative pose estimation on the KITTI odometry dataset. We have observed that with the pairwise matching supervision, the result for motion estimation has been extensively improved. We measure the Absolute Trajectory Error (ATE) over $N$-frame snippets. The mean error and variance are averaged from the full sequence. As shown in Table~\ref{tab:pose}, with the same underlying network structure, the proposed method outperforms state-of-the-art methods by a large margin. 

However, the above comparison is in favor of learning-based approaches which are only able to generate $N$-view pose segments, but not fair for mature SLAM systems which emphasizes the accuracy of the full trajectory.
To demonstrate that our method produces consistent pose estimation, we chain the relative poses by averaging the two overlapping frames of 3-view snippets. We first align the chained motion with the ground-truth trajectory by estimating a similarity transformation~\cite{umeyama1991least}, and then compute the average of APE for each frame.
As shown in Figure~\ref{fig:slam_traj}, even without global motion averaging techniques~\cite{govindu2006robustness}, our method achieves comparable performance (8.82m/23.09m median APE for Seq. 9/10) against monocular ORB-SLAM~\cite{mur2015orb} (36.83m/5.74m median APE for Seq. 9/Seq. 10) without loop closure. This comparison just provides a fair comparison in terms of the full sequence, yet by no means shows the learning-based method has surpassed tradition VO methods. In fact, monocular ORB-SLAM with loop closure and global bundle adjustment results in a much smaller 7.08m median APE for Seq. 9 (Seq. 10 stays unchanged because it has no loop).

\subsection{Ablation Study}
	\begin{table*}[]
		\resizebox{\textwidth}{!}{ 
			\begin{tabular}{c|c|c|c|c|c||c|c|c|c|c|c|c|c|c}
				\hlineB{3}
				\multicolumn{6}{c||}{\textbf{Loss Configuration}}                                                       & \multicolumn{7}{c|}{\textbf{Depth} (KITTI raw Eigen split)}                                                                                                                                                                        & \multicolumn{2}{c}{\textbf{Pose} (KITTI odometry)}                        \\ \hline
				Baseline & Epipolar & Re-projection & Forward-backward & Multi-view & Mask & \multicolumn{1}{c|}{\cellcolor{red!20}Abs Rel} & \multicolumn{1}{c|}{\cellcolor{red!20}Sq Rel} & \multicolumn{1}{c|}{\cellcolor{red!20}RMSE} & \multicolumn{1}{c|}{\cellcolor{red!20}RMSE log} & \multicolumn{1}{c|}{\cellcolor{green!25} $\delta < 1.25^3$} & \multicolumn{1}{c|}{\cellcolor{green!25} $\delta < 1.25^2$} & \multicolumn{1}{c|}{\cellcolor{green!25} $\delta < 1.25^3$} & \multicolumn{1}{c|}{Seq 09} & \multicolumn{1}{c}{Seq 10} \\ \hlineB{3}
				$\checkmark$                      & -             & -                  & -                & -          & -         &     0.163                         &     1.371                        &   6.275                        &      0.249                         &         0.773                  &  0.918                         &     0.966                      &         0.014 $\pm$ 0.009                    &      0.012 $\pm$ 0.012                       \\ 
				$\checkmark$                      & $\checkmark$             & -                  & -                & -          & -         &     0.159                         &   1.287                          &  5.725                         &    0.239                           &   0.791                        & 0.927                          &   0.969                        &    0.010 $\pm$ 0.005                          &       0.009 $\pm$ 0.008                       \\ 
				$\checkmark$                      & $\checkmark$             & $\checkmark$                  & -                & -          & -         &          0.152                    &     1.205                        &      5.56                     &     0.227                          &      0.800                     &      0.935                     &       0.973                    &     0.009 $\pm$ 0.005                        &       0.009 $\pm$ 0.008                      \\ 
				$\checkmark$                     & $\checkmark$             & $\checkmark$                  & $\checkmark$                & -          & -         &            0.146                  &    1.391                         &         5.791                  &      0.229                         &          0.814                 &     0.936                      &       0.972                    &      0.009 $\pm$ 0.005                       &   0.008 $\pm$ 0.007                          \\ 
				$\checkmark$                     & $\checkmark$             & $\checkmark$                  & $\checkmark$               & $\checkmark$          & -         &          0.143                    &    1.114                         &       5.681                    &     0.225                          &      0.816                     &      0.938                     &       0.974                    &    \textbf{0.009 $\pm$ 0.005}                         &        \textbf{0.008 $\pm$ 0.007}                    \\ 
				$\checkmark$                    & $\checkmark$             & $\checkmark$                 & $\checkmark$               & $\checkmark$         & $\checkmark$         &       \textbf{0.140}                       &    \textbf{ 1.025}                        &   \textbf{5.394}                        &     \textbf{0.222}                          &    \textbf{0.816}                       &     \textbf{0.938}                      &    \textbf{0.974}                       &       \textbf{0.009 $\pm$ 0.005}                       &       \textbf{0.008 $\pm$ 0.007}                       \\ \hlineB{3}
				$\checkmark$ (5-view)                     &  -           & -                  & -               &      -     &     -     &       0.169                       &     1.607                       &   6.129                        &     0.255                          &    0.779                       &     0.917                      &    0.963                       &       0.014 $\pm$ 0.009                       &       0.013 $\pm$ 0.009                       \\
				
				$\checkmark$ (5-view)                     &  $\checkmark$          & $\checkmark$                & -               &      -     &     -     &       0.157                       &    1.449                       &   5.796                        &     0.239                         &    0.803                       &     0.929                     &    0.970                       &       0.012 $\pm$ 0.008                       &       0.010 $\pm$ 0.007                       \\ \hlineB{3}
			\end{tabular}
		}
\caption{Evaluation of different training loss configurations. All models are either solely trained on KITTI raw dataset (for depth) or KITTI odometry dataset (for pose) without pre-training on Cityscapes. The depth estimation performance is evaluated with maximum depth set/capped at 80m. All models except the last two are trained on 3-view image sequences. The best result for each metric is \textbf{bolded}.}
	\label{tab:diff_mod}
\end{table*}

\noindent\textbf{Performance with different modules.} We conduct an ablation study to show the effect of each component. The models for depth and pose evaluation are trained solely on KITTI raw dataset and odometry dataset respectively. 
We choose an incremental order for the proposed techniques to avoid too many loss term combinations. 
As shown in Table~\ref{tab:diff_mod}, we have the following observations:
\begin{itemize}[leftmargin=*]
	\setlength{\parskip}{2pt}
	\setlength{\itemsep}{0pt plus 1pt}
	\item The re-implemented baseline model, using Equation~\ref{eqn:base_loss}, has already surpassed several models~\cite{klodt2018supervising,mahjourian2018unsupervised,zhou2017unsupervised}. The reasons can be attributed to the more capable depth encoder ResNet-50, which is also used by~\cite{yin2018geonet}.
	\item The result for pose estimation is greatly improved with the epipolar loss term $\mathcal{L}_{epi}$. It shows the efficacy of using raw feature matches as the weakly supervised signal. However, the improvement for depth estimation is not as significant as pose estimation.
	\item Re-projection loss further improves the depth inference. The improvements for pose estimation brought by ingredients other than the epipolar loss are marginal.
	\item The depth consistency and multi-view consistency are the essential parts for the improvement in depth estimation.
\end{itemize}
In summary, the epipolar geometric supervision helps the pose estimation most, while the geometric consistency terms in Section~\ref{sec:depth_est} essentially improve depth estimation.

\smallskip
\noindent\textbf{Sequence Length.} The multi-view depth consistency loss boosts the depth estimation. However, the performance boost can be also attributed to using longer image snippets, since similar second-order relations can be exploited by using 5-view image sequences for training. Therefore, we further evaluate the performance of using 5-view images. As shown in Table~\ref{tab:diff_mod}, training on longer image sequences would deteriorate the performance, because long sequences also contain larger photometric noises. It shows that the proposed formulation elevates the results not from more data, but the consistency embedded in geometric relations.

\subsection{Generalization on Make3D}
\begin{table}[]
	\resizebox{0.48\textwidth}{!}{ 
	\begin{tabular}{l|c|c|c|c|c|c}
		\hlineB{3}
		\multirow{2}{*}{Method} & \multicolumn{2}{c|}{Supervision} & \multicolumn{4}{c}{Metrics}        \\ \cline{2-7} 
		& depth           & pose           & \cellcolor{red!20}Abs Rel & \cellcolor{red!20}Sq Rel & \cellcolor{red!20}RMSE  & \cellcolor{red!20} RMSE $\log_{10}$ \\ \hlineB{3}
		Karsch \etal~\cite{karsch2014depth}\dag           &     $\checkmark$
		            &        -        & 0.417   & 4.894  & 8.172 & 0.144    \\ 
		Liu \etal~\cite{liu2014discrete}\dag             &         $\checkmark$        &      -          & 0.462   & 6.625  & 9.972 & 0.161    \\ 
		Laina \etal~\cite{laina2016deeper} \dag           &       $\checkmark$          &       -         & 0.198   & 1.665  & 5.461 & 0.082    \\ 
		Godard \etal~\cite{godard2017unsupervised}           &      -           &       $\checkmark$         & 0.443   & 7.112  & 8.860 & 0.142    \\
		Zhou \etal~\cite{zhou2017unsupervised}           &      -           &       -         & 0.392   & 4.473  & 8.307 & 0.194    \\
		Ours                &         -        &      -          & 0.378   & 4.348  & 7.901 & 0.183        \\ \hlineB{3}
	\end{tabular}
		}
\caption{Generalization experiments on Make3D. The evaluation metrics are the same as the ones in Table~\ref{tab:depth_est} except the last one (RMSE $\log_{10}$) to conform with~\cite{karsch2014depth}. The methods marked with $\dag$ are trained on Make3D. The depth estimation is evaluated with maximum depth capped at 70m. We use the center-cropped images as in~\cite{godard2017unsupervised} and resize them to $128 \times 416$ for inference.}
\label{tab:make3d}
\end{table}

\begin{figure}[th]
\resizebox{0.48\textwidth}{!}{ 
	\centering 
	\includegraphics[width=\textwidth]{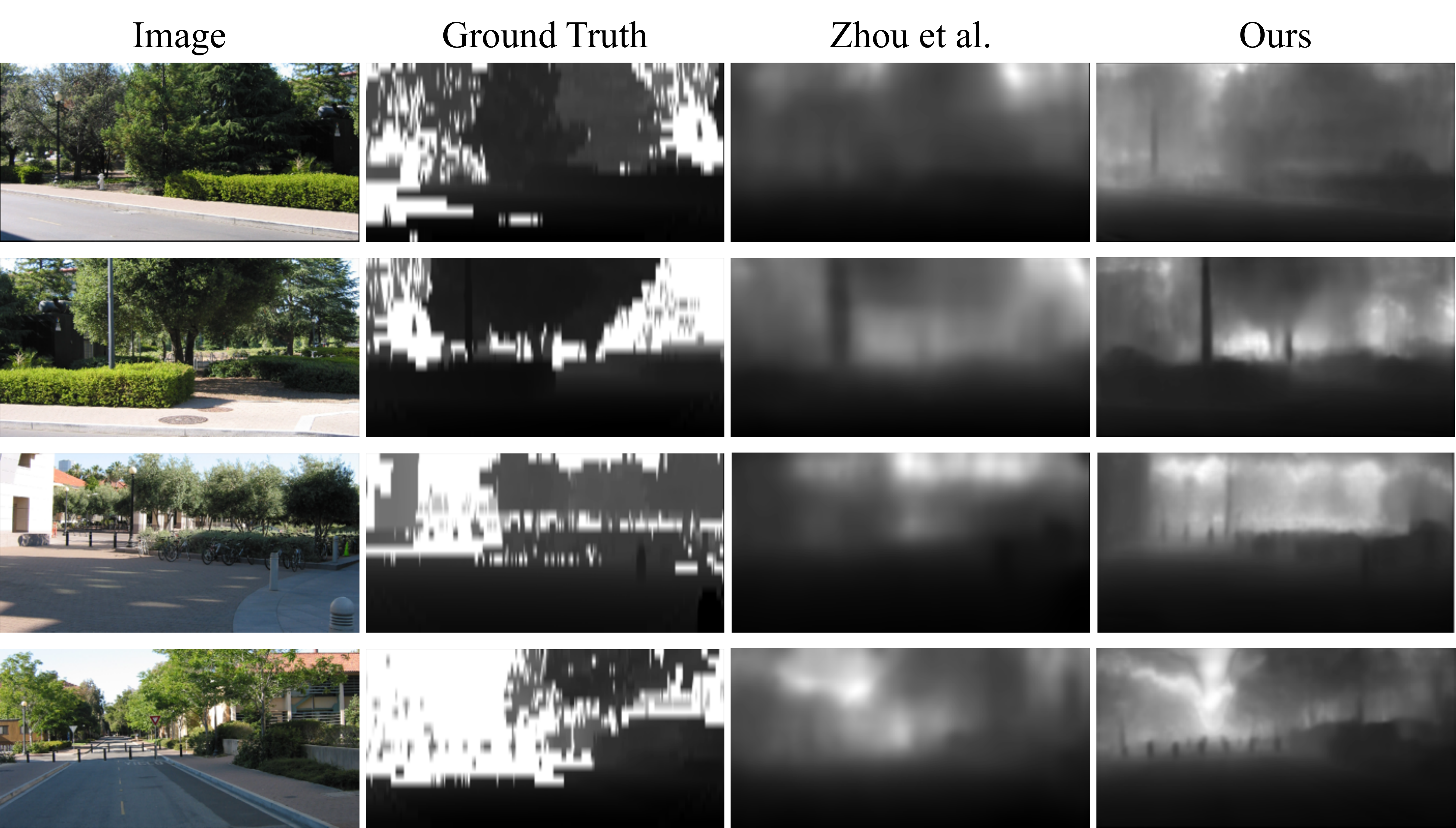}
}
	\caption{Sample depth predictions on the Make3D dataset. Both our method and SfMlearner~\cite{zhou2017unsupervised} are trained on Cityscapes+KITTI.}
	\label{fig:make3d}
\end{figure}

To illustrate that the proposed method is able to generalize to other datasets unseen in the training, we compare to several supervised/self-supervised methods on the Make3D dataset~\cite{saxena2009make3d}, using the same evaluation protocol as in~\cite{godard2017unsupervised}.
As shown in Table~\ref{tab:make3d}, our best model achieves reasonable generalization ability and even beats several supervised methods on some metrics. A qualitative comparison is shown in Figure~\ref{fig:make3d}.

\section{Conclusion}
We have presented an unsupervised pose and depth estimation pipeline that absorbs both the geometric principles and learning-based metrics. We emphasize on the consistency issue and propose novel ingredients to make the result more robust and reliable. Yet, we should realize that the current learning-based methods are still far from solving the SfM problem in an end-to-end fashion. Further investigations include enforcing consistency across the whole dataset, such as incorporating loop closure and bundle adjustment techniques into the learning-based methods.

\clearpage
{\small
\bibliographystyle{ieee_fullname}
\bibliography{egbib}
}

\clearpage

\section{Appendix}
\subsection{Triangulation Uncertainty}
As shown in figure~\ref{fig:forward_motion}, the case with forward motion, which is often encountered in the driving scenarios, renders high uncertainty with the triangulated 3-d track. To model its uncertainty, we consider the simplified case where the point on a plane has 2-d position $\mathbf{X}$ given it two 1-d image points $p = f(\mathbf{X}) = P_{2\times 3} \mathbf{X}$ and $p' = f'(\mathbf{X}) = P_{2\times 3}' \mathbf{X}$ using line cameras~\cite{hartley2003multiple}. Line camera projects plane points to line points, which is similar to the normal pinhole camera model that projects 3-d points to 2-d points. Suppose the measurement $\hat{p}$ and $\hat{p}'$ are corrupted by the Gaussian noise $\mathcal{N}(0, \sigma^2)$, then the probability of obtaining $\hat{p}$ given the 2-d point $\mathbf{X}$ is 
\begin{equation}
\text{Prob}(p | \mathbf{X}) = \frac{1}{\sqrt{2\pi \sigma^2}} e ^{-\frac{(p-f(\mathbf{X}))^2}{2\sigma^2}}
\end{equation}
We assume $p'$ has the same probability, then the posterior probability of obtaining $\mathbf{X}$ given $p$ and $p'$ is
\begin{equation}
\begin{split}
\text{Prob}(\mathbf{X} | p, p') &= \frac{\text{Prob}(p, p' | \mathbf{X}) \text{Prob}(\mathbf{X})}{\text{Prob}(p, p')}\\
&\sim \text{Prob}(p | \mathbf{X}) \text{Prob}(p' | \mathbf{X})
\end{split}
\end{equation}
assuming a uniform prior distribution for $\mathbf{X}$ and the two measurements $p$ and $p'$ are independent identically distributed (i.i.d.). The bias and variance of this distribution is illustrated with the shaded area in Figure~\ref{fig:forward_motion} intuitively using the angle between the rays, and is also
discussed in~\cite{hartley2003multiple}. The variance of the triangulated point would be high if the camera possesses the forward motion. 

Though using the triangulated point as the depth supervision is also feasible, we argue that this is inferior to using re-projection error because 1) the depth of the triangulated track is not stable thus may incur large gradient for distant tracks; 2) even if the depth predictions for distant tracks are wrong, it would not incur much error for the gradient and therefore make the training more stable; 3) the triangulation computation itself imposes additional computation in the training, as this process involves the relative pose and should be conducted online.

\subsection{Full Pose Prediction on KITTI}
The accurate pose prediction is one of the significant improvements brought by the proposed method, which is not feasible for supervised approaches~\cite{eigen2014depth,liu2016learning} nor semi-supervised approaches that rely on stereo pairs~\cite{garg2016unsupervised,godard2017unsupervised}. Here we show more pose prediction comparisons with monocular ORB-SLAM~\cite{mur2015orb}, one of the state-of-the-art indirect SLAM methods. We remove the loop closure functionality from the original ORB-SLAM (denoted as \textit{M-ORB-NO-CL}) to conduct a fair comparison since it is currently infeasible to apply such technique to the learning-based methods. For the proposed method, we simply average the rotation and translation of the adjacent 3-view pose predictions to obtain the full pose trajectory, fixing the first frame as the origin and incrementally aligning the 3-view poses. This simple alignment without global motion averaging~\cite{govindu2006robustness} may incur biases with respect to the starting point, but we would like to emphasize the consistency and potential of the proposed method. We evaluate the median absolute position error (in meters) on KITTI odometry sequence 00-10.

\begin{table*}[]
	\resizebox{\textwidth}{!}{ 
		\begin{tabular}{l|ccccccccccc}
			\hlineB{3}
			& \cellcolor{red}\textit{Seq. 00} & \cellcolor{red}\textit{Seq. 01} & \cellcolor{red}\textit{Seq. 02} & \cellcolor{red}\textit{Seq. 03} & \cellcolor{red}\textit{Seq. 04} & \cellcolor{red}\textit{Seq. 05} & \cellcolor{red}\textit{Seq. 06} & \cellcolor{red}\textit{Seq. 07} & \cellcolor{red}\textit{Seq. 08} & \cellcolor{green}\textit{Seq. 09} & \cellcolor{green}\textit{Seq. 10} \\ \hlineB{3}
			\#Frame     & 4541             & 1101             & 4661             & 801              & 271              & 2761             & 1101             & 1101             & 4071             & 1591             & 1201             \\ \hline
			\textit{Ours mAPE (m)}        & 38.27            & \textbf{96.62}   & \textbf{48.18}   & 7.19             & 1.79             & \textbf{9.82}    & \textbf{6.97}    & \textbf{4.82}    & \textbf{24.11}   & \textbf{8.82}    & 23.09            \\
			\textit{M-ORB-NO-CL mAPE (m)} & \textbf{17.76}   & F                & 58.27            & \textbf{0.48}    & \textbf{0.39}    & 31.24            & 47.17            & 13.16            & 36.37            & 39.86            & \textbf{4.57}    \\\hline
			\textit{Ours time (s)}        & 8.18            & 3.56   & 8.08   & 3.25             & 2.57             & 5.67    & 3.64    & 3.61   & 7.27   & 4.20    & 3.76            \\ 
			\textit{M-ORB-NO-CL time (s)} & 481.83   & 122.30                & 493.95            & 90.59    & 35.18    & 295.41            & 121.89           & 122.20            & 431.92            & 173.14            & 132.71    \\ \hlineB{3}
		\end{tabular}
	}
	\caption{The median absolute position error (mAPE) and running time of different methods for all KITTI odometry sequences with ground-truth (00-10). `F' means fail. The better results are \textbf{bolded}. The columns marked by {\color{red}red} are training sequences, and those marked by {\color{green}green} are originally used for testing.}
	\label{tab:full_kitti}
\end{table*}

\begin{figure*}[t]
	\centering 
	\includegraphics[width=\textwidth]{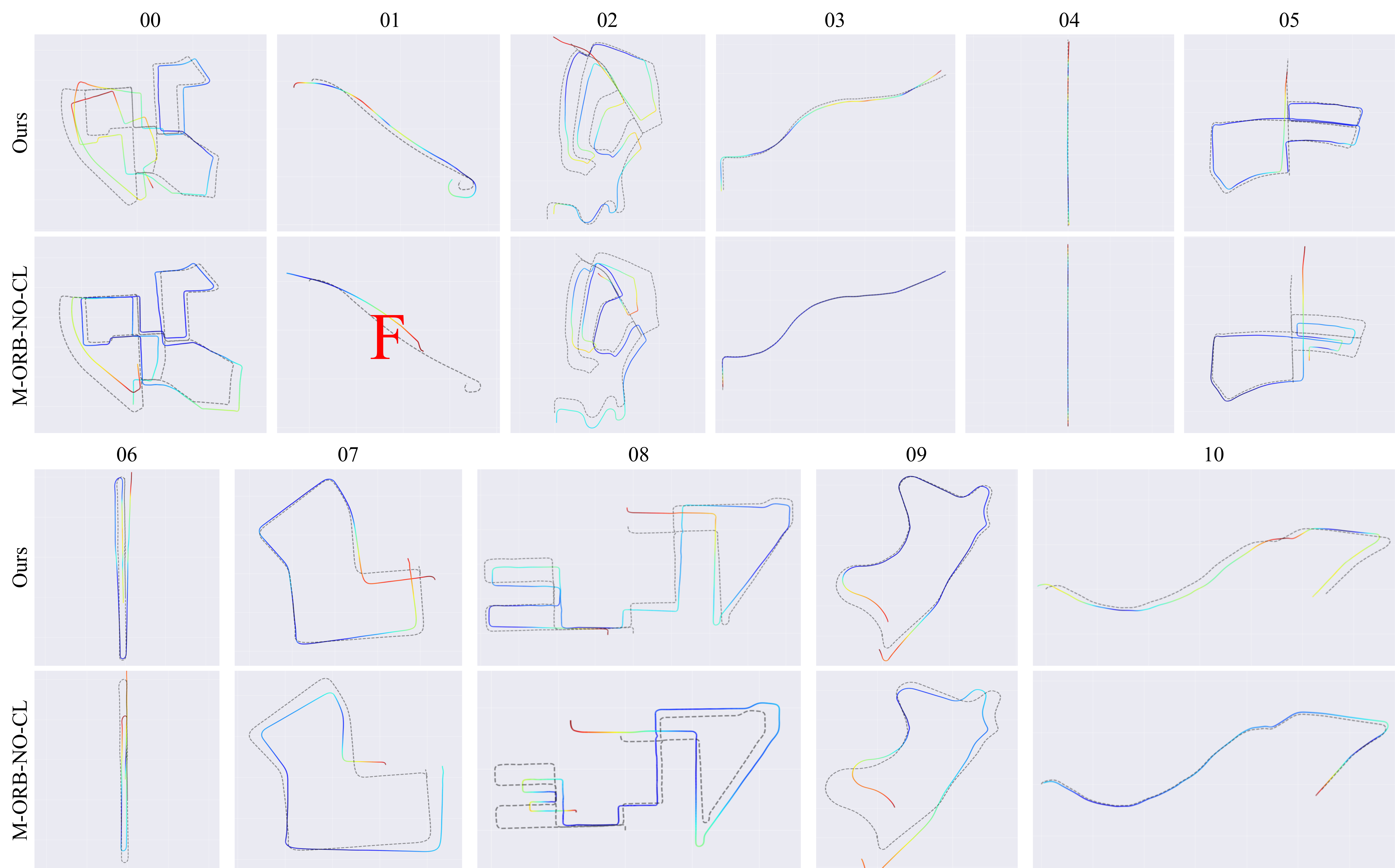}
	\caption{Comparisons of full trajectories for KITTI odometry sequence 00-10. {\color{red}F} means the method is failed on this sequence. The dotted line in each sub-figure is the ground-truth trajectory.}
	\label{fig:kitti_all}
\end{figure*}

As shown in Table~\ref{tab:full_kitti}, the proposed method achieves better performance on 7 of the 11 sequences. In addition, the proposed method is an order of magnitude faster than \textit{M-ORB-NO-CL}. It is noted that the proposed method utilizes GPU (GTX 1080 Ti) while ORB-SLAM is run on CPU (Intel Core i7-4770k), yet the traditional methods are hard to compete with the learning-based methods in terms of efficiency.
Monocular ORB-SLAM would fail on \textit{Seq. 01}, the scene of a car moving on a high-way, which is difficult for sparse feature detection and matching. In this case, learning-based methods have the advantage that a coarse result is guaranteed, even if it is inaccurate. Note that monocular ORB-SLAM usually takes the first a few frames for initialization (which we set to $[\mathbf{I}|\textbf{0}]$), while learning-based methods do not need this initialization step. Figure~\ref{fig:kitti_all} shows the side-by-side comparison of the full trajectories.
\end{document}